\documentclass{article}
\usepackage{amsmath,epsfig}
\usepackage[a4paper, left=2cm, right=2cm]{geometry}

\begin{document}

\begin{center}
{\Large Power Plant Classification from Remote Imaging with Deep Learning}\\
\vspace{1em}
\textbf{Michael Mommert, Linus Scheibenreif, Jo\"{e}lle Hanna, Damian Borth}\\
University of St.\ Gallen, Institute of Computer Science\\
\vspace{2em}
Presented at the 2021 IEEE International Geoscience and Remote Sensing Symposium (IGARSS).
\end{center}
%
%
%

%
%
\begin{abstract}
  Satellite remote imaging enables the detailed study of land use
  patterns on a global scale. We investigate the
  possibility to improve the information content of traditional land
  use classification by identifying the nature of industrial sites
  from medium-resolution remote sensing images. In this work, we focus
  on classifying different types of power plants from Sentinel-2
  imaging data. Using a ResNet-50 deep learning model, we are able to
  achieve a mean accuracy of 90.0\% in distinguishing 10 different
  power plant types and a background class. Furthermore, we are able
  to identify the cooling mechanisms utilized in thermal power plants
  with a mean accuracy of 87.5\%. Our results enable us to
  qualitatively investigate the energy mix from Sentinel-2 imaging
  data, and prove the feasibility to classify industrial
  sites on a global scale from freely available satellite imagery.
\end{abstract}


%
\section{Introduction}
\label{sec:intro}

The availability of satellite remote imaging data with global
coverage and high imaging cadence enable the
detailed investigation of land use patterns and changes therein. Land
use classification creates a mapping of how land is
being utilized by humans; such information is highly valuable to
better understand socio-economic, environmental, and ecological
problems and to inform policy-makers. However, the value of land use
information is limited by both the spatial and categorical resolution
of the data, the latter referring to the level of detail utilized in
the classification or segmentation.

We investigate the use of deep learning applied to satellite remote
imaging to identify the nature of industrial sites in an attempt to
improve the level of detail of land use classification maps. Previous
works have successfully combined both approaches for land use
classification \cite{Zhang2018}, to detect solar \cite{Yu2018} and
wind power facilities \cite{Zhou2019}, as well as oil and gas
processing facilities \cite{Sheng2020}. However, all these works rely
on high-resolution remote imaging data, which are typically associated
with high cost and have a low imaging cadence.  In this work, we
investigate the use of freely available medium-resolution and
high-cadence satellite imaging data to classify different types of
power plants as a test bed for a future application to industrial
sites. This application also addresses the topic of climate change
mitigation by enabling the qualitative analysis of the energy mix from
satellite imagery.

\section{Data Set}
\label{sec:dataset}

For this study, we select a total of 450 different power plant sites
from the JRC Open Power Plants Database (JRC-PPDB-OPEN, \cite{JRC}),
which in addition to general power plant type information also
provides approximate geolocation data for each site. For each power
plant class, we extract up to 50 different sites for which we
verify and improve
geolocation data based on high-resolution remote imaging data provided
by the Google Maps service. Our resulting sample contains the
following power plant classes (nomenclature as used in JRC-PPDB-OPEN,
number in parentheses refer to the number of different
sites in the sample):
``Fossil Brown coal/Lignite'' (50), ``Fossil Gas'' (50), ``Fossil Hard
Coal'' (50), ``Fossil Oil'' (30), ``Hydro Pumped Storage'' (50),
``Hydro Run-of-river and poundage'' (50), ``Hydro Water Reservoir''
(50), ``Nuclear'' (50), ``Solar'' (20), ``Wind Onshore'' (50).

Images utilized in this work were acquired with the Multi-Spectral
Imagers (MSI) onboard the two Sentinel-2 Earth-observing
satellites operated by the European Space Agency. Both
satellites observe the land surface of the Earth in 13 bands covering
the optical to short-wave infrared wavelengths at spatial resolutions
of 10~m to 60~m per pixel and with an imaging cadence of
${\geq}$5~days for most locations on Earth.

For each site, we download a total of 10 different and mostly
cloud-free Sentinel-2 rasters taken during 2020 to ensure proper
coverage of seasonal and other temporal changes in each site.  Around
each location, we crop a square image region with an edge length of
1~km. We utilize only those MSI imaging channels that do not focus on
atmospheric features (bands 2, 3, 4, 5, 6, 7, 8, 8A, 11, and 12) and
upsample all bands to the highest image resolution (10~m/pixel).  We
perform the same extraction procedure on 4 randomly selected locations
on each raster, resulting in our background images, which are used as
a control group to quantify the detectability of power plants against
random settings.

Our data set consists of 4,154 images of 450 different power plant
sites across Europe and 16,064 random background images. Example
images are shown in Figure \ref{fig:examples}.

\section{Methods}
\label{sec:methods}

Deep Neural Network approaches are well-suited to deal with the
complexity inherent to remote imaging data. We approach the
multi-class classification problems in this work with a ResNet-50 deep
residual neural network \cite{He2016} that was slightly modified: the
first convolutional layer takes in 10-channel input images (see
Section \ref{sec:dataset}) and the final fully connected layer returns
a number of logits that is equal to the number of classes of the
respective problem.  Furthermore, the first convolutional and maxpool
layers have been modified to use a 3$\times$3 kernel size and a stride
of unity to enforce sensitivity to small-scale structure in our
images.

For all experiments we utilize stochastic gradient descent with
momentum as our optimizer. The learning is based on a cross entropy
loss function.
Our deep learning methods are implemented utilizing
PyTorch \cite{Paszke2019}.

To boost the learning and results of our classifiers, we pretrain our
models on the EuroSAT dataset \cite{Helber2019}, which contains a
sample of Sentinel-2 image patches with corresponding land use
classification labels.

We randomly split our data set on a per-image basis into a training,
validation, and test data set using a 80\%/10\%/10\% split. Before
provided to the model for either training or evaluation, each image is
normalized based on the mean and standard deviation of each
Sentinel-2/MSI band across all images. Furthermore, random image
flipping, mirroring, and rotations by integer multiples of 90 degrees
are applied to provide data augmentation. For each epoch during
training, validation, and testing, balanced samples are drawn from the
corresponding data sets, using either subsampling or oversampling
depending on the number of available images in each class.

\section{Results}
\label{sec:results}

\subsection{Power Plant Classification}
\label{sec:results_pp}

\begin{figure}[h]
  \centering
  \includegraphics[width=\linewidth]{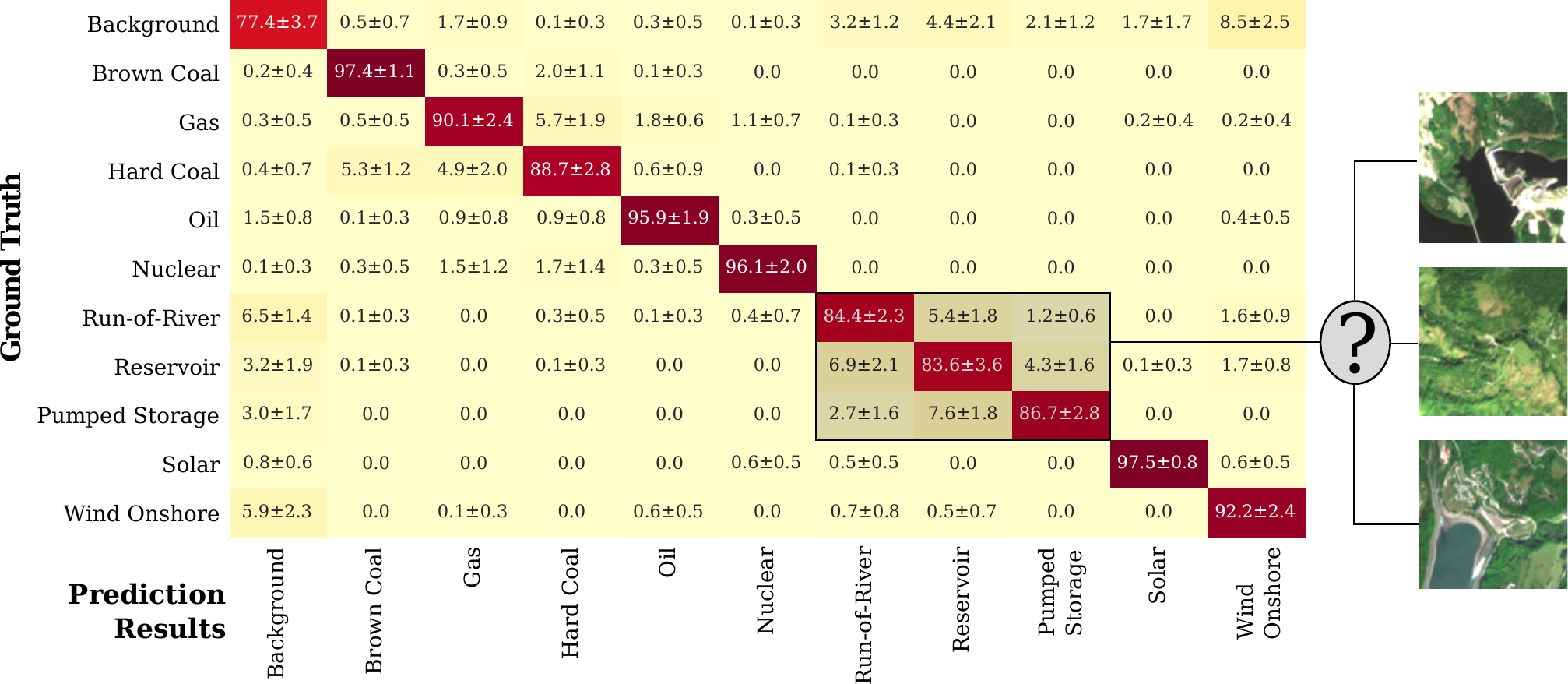}
  \caption{Confusion matrix for the classification of 10 different
    power plant classes (classes ``Brown Coal'', ``Gas'', ``Hard
    Coal'', and ``Oil'' refer to fossil fuel power plants, whereas
    ``Run-of-River'', ``Reservoir'', and ``Pumped Storage'' refer to
    hydroelectric plants) and one background class. Values are
    accuracies expressed as percentages. Uncertainties, derived as
    standard deviations across ten different randomly drawn balanced
    samples from the test data set, are quoted where significant.
    While most classes have high individual detection accuracies,
    there are chances of confusion among the different hydroelectric
    classes (see example images on the right) and the fossil fuel
    plant classes, as well as between the
    renewable energy classes and the background class.
  \label{fig:ppclassification}}
\end{figure}

We train our modified ResNet-50 architecture on the classification
task involving 10 different power plant classes (see Section
\ref{sec:dataset}) and a background class. After tuning the
hyperparameters of the model on the validation data set, we evaluate
the model performance on the test data set. We find an overall
accuracy over all classes of 90.0\%; the accuracy of individual classes
varies between 77.4\% for the background class and 97.5\% for solar
power plants. Figure \ref{fig:ppclassification} shows the per-class
accuracies and corresponding uncertainties in detail. The accuracy
variations between power plant types correlate well with the presence
or absence of unique feature that would enable a clear identification.

\subsection{Power Plant Cooling Classification}

Thermal power plants utilize different methods of cooling necessary
for the mechanical generation of power from steam. As provided by
JRC-PPDB-OPEN, our full data set contains 4 different cooling types
among the thermal power plants (all fossil fuel plants and nuclear
plants): air cooling (29 examples), mechanical draft tower (22
examples), natural draft tower (84 examples), once-through cooling (90
examples). We ignore 5 examples of thermal power plants that do not
utilize active or passive cooling.  The full data set of thermal power
plants includes 2,112 images.

We investigate the possibility to also classify the type of cooling
used with the same ResNet-50 architecture as outlined in Section
\ref{sec:methods}. Utilizing the same approach as introduced above, we
train the model on the reduced data set of thermal power plants and
train against the 4 different target classes (see above). We utilize
the trained model from Section \ref{sec:results_pp} as a pretrained
model for this task.

Our trained model achieves a mean accuracy of 87.5\% over all cooling
classes. The individual class accuracies range from 70.0\% for
mechanical draft towers to 99\% for once-through cooling. This range
is easily explainable as once-through cooling systems require access
to a water body that is readily detectable, whereas mechanic draft
towers can resemble other buildings and structures.

\section{Discussion and Conclusions}
\label{sec:discussion}

\begin{figure}[t]
  \centering
  \includegraphics[width=0.35\linewidth]{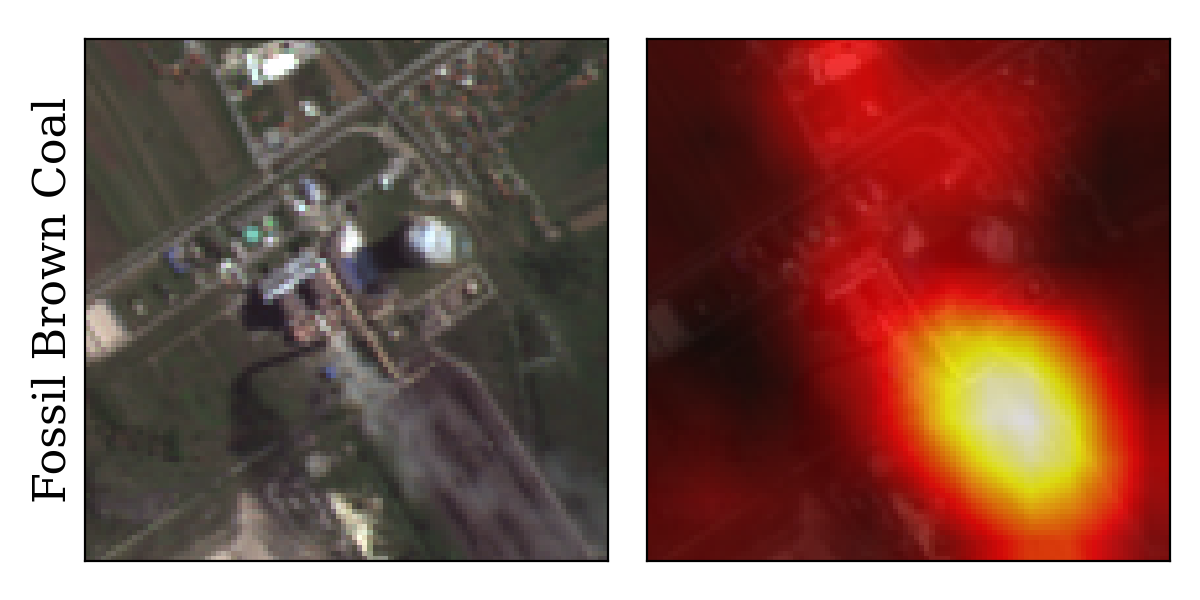}
  \hspace{5em}
  \includegraphics[width=0.35\linewidth]{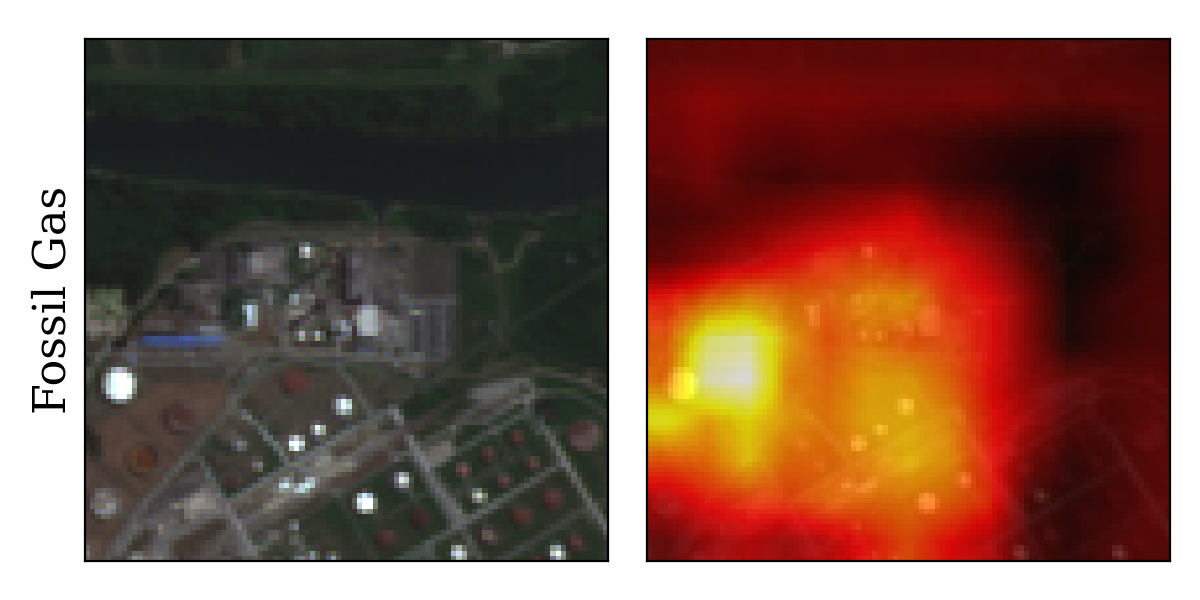}
  \includegraphics[width=0.35\linewidth]{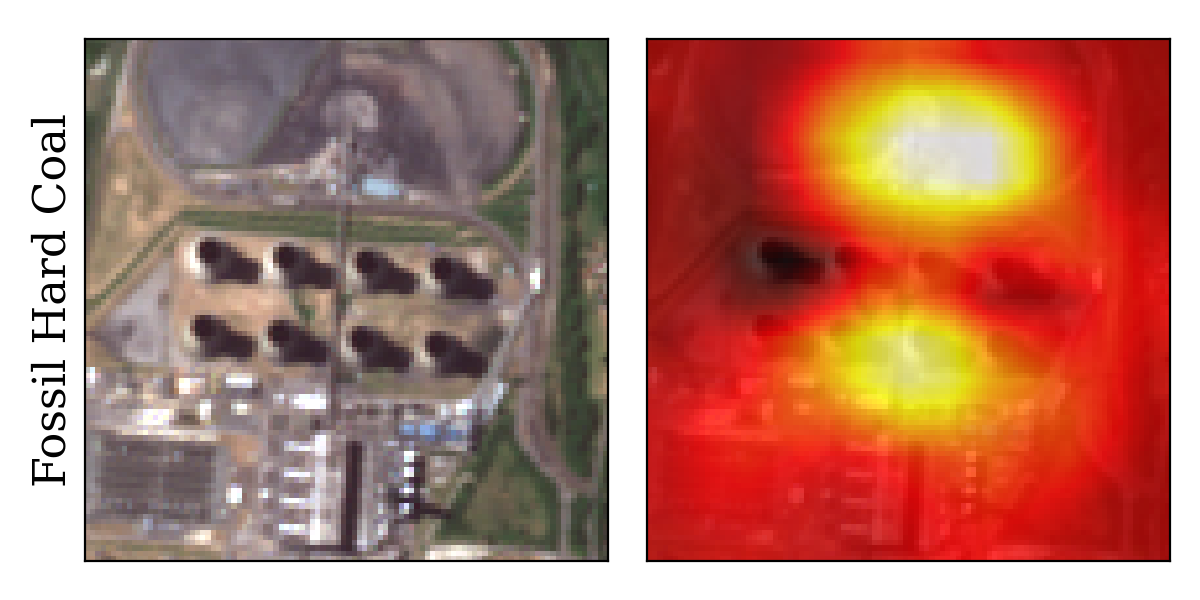}
  \hspace{5em}
  \includegraphics[width=0.35\linewidth]{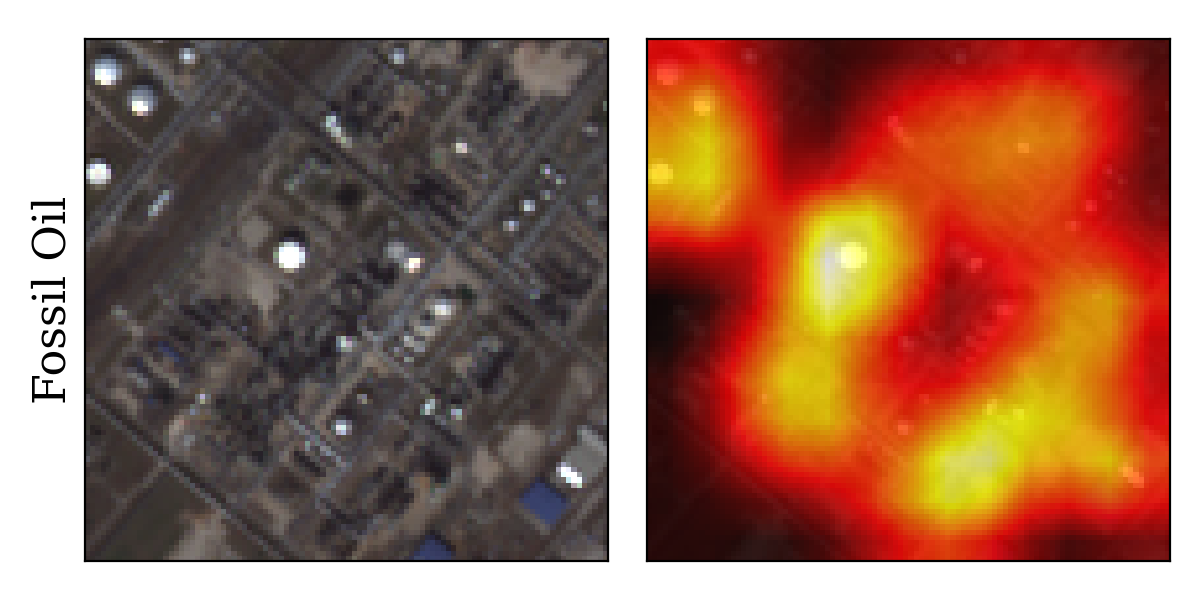}
  \includegraphics[width=0.35\linewidth]{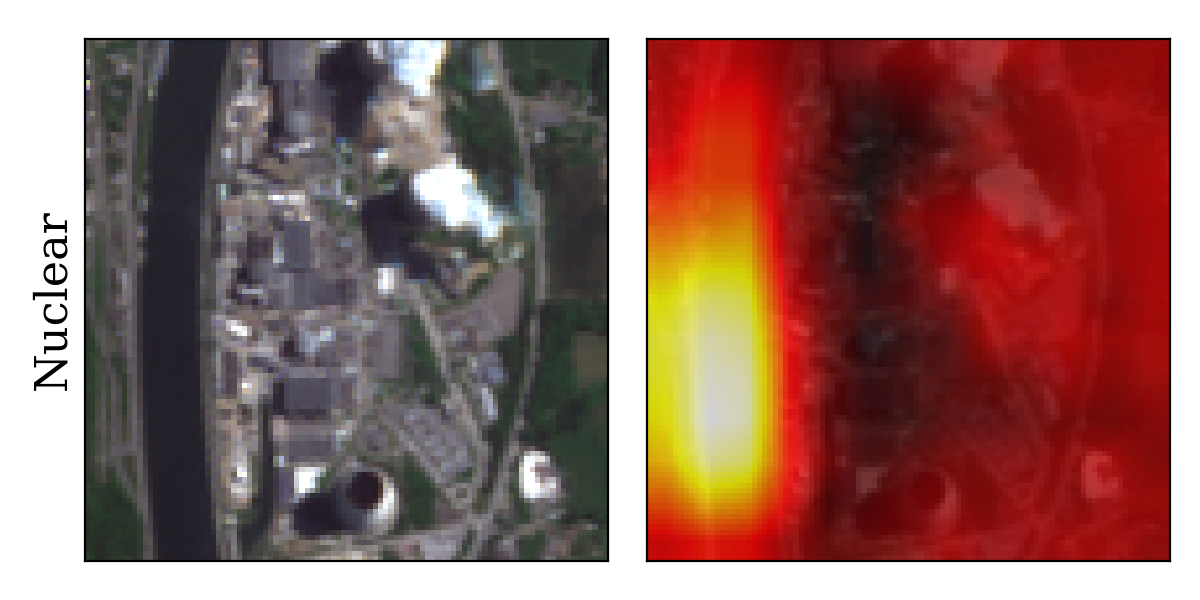}
  \hspace{5em}
  \includegraphics[width=0.35\linewidth]{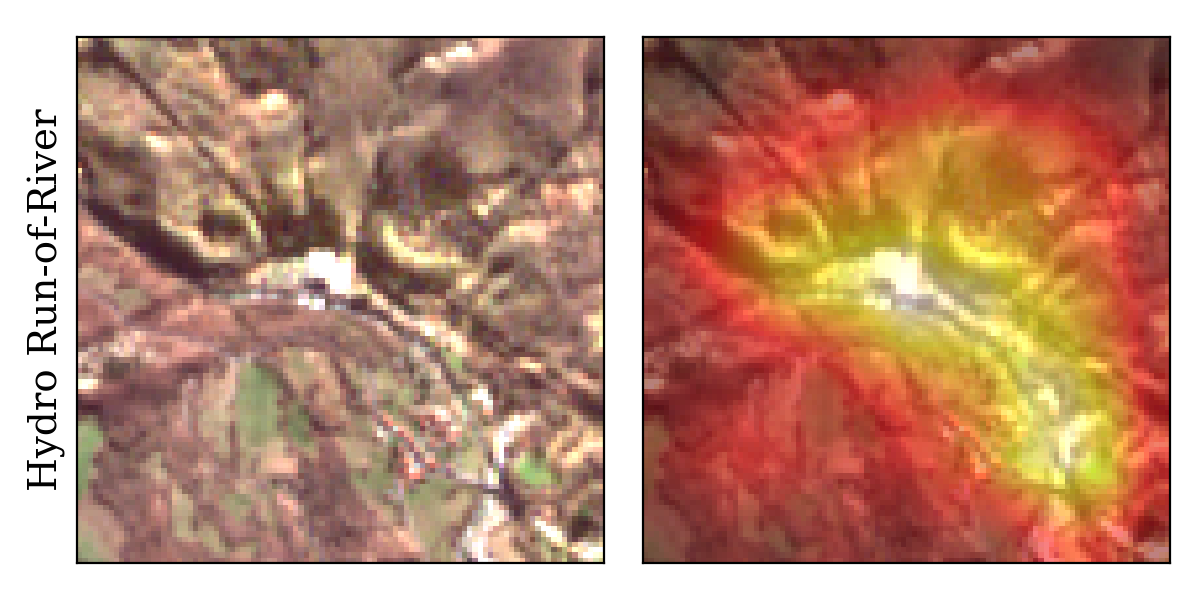}
  \includegraphics[width=0.35\linewidth]{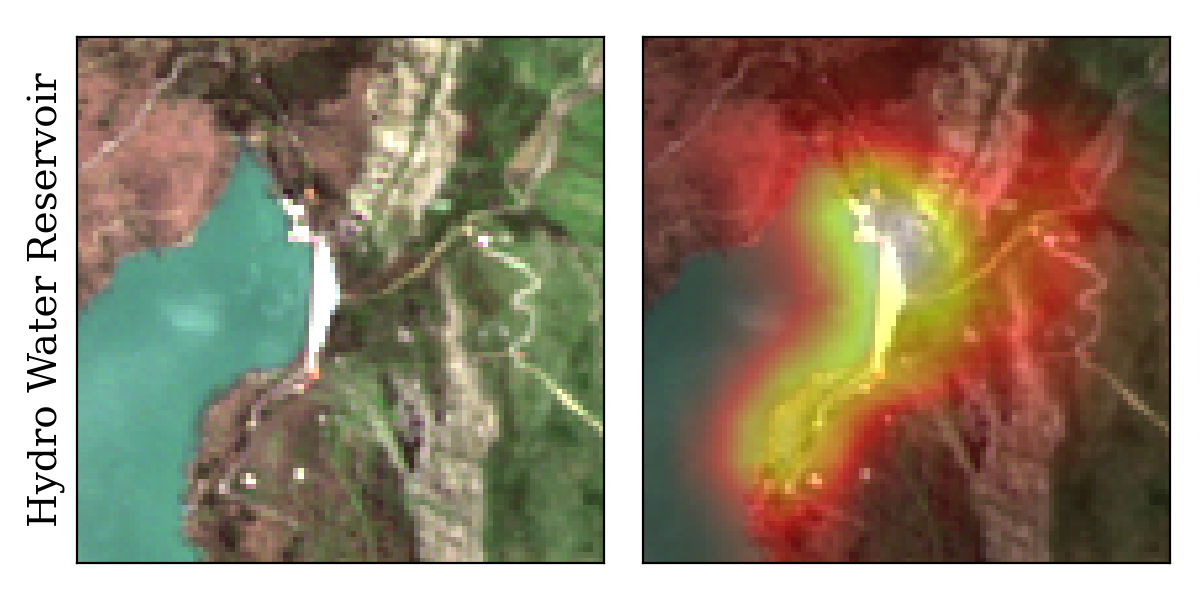}
  \hspace{5em}
  \includegraphics[width=0.35\linewidth]{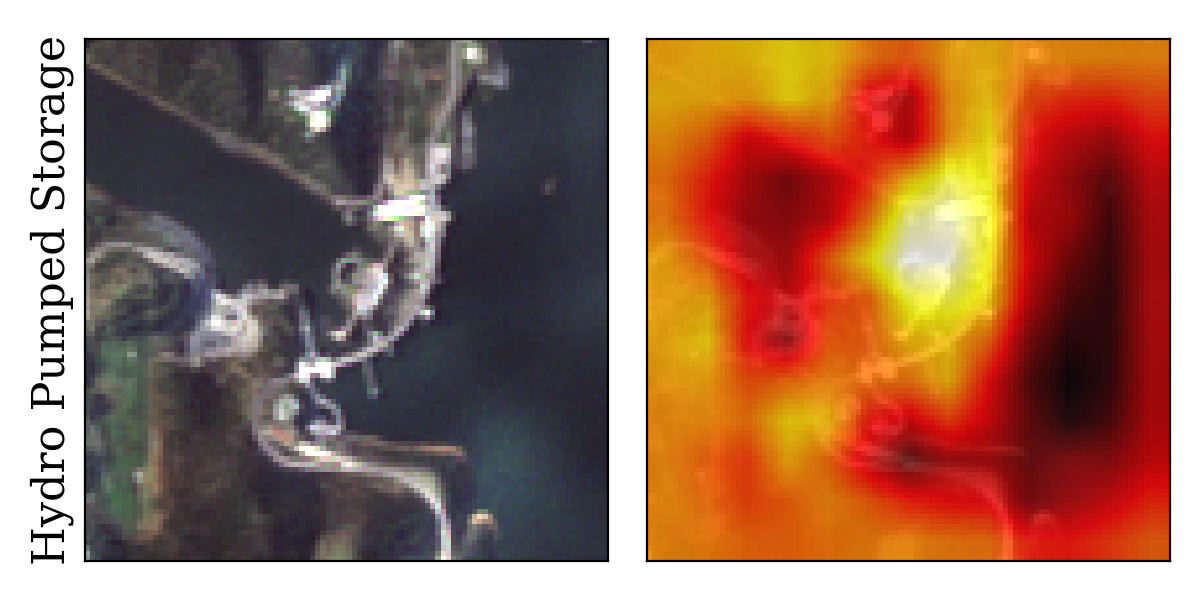}
  \includegraphics[width=0.35\linewidth]{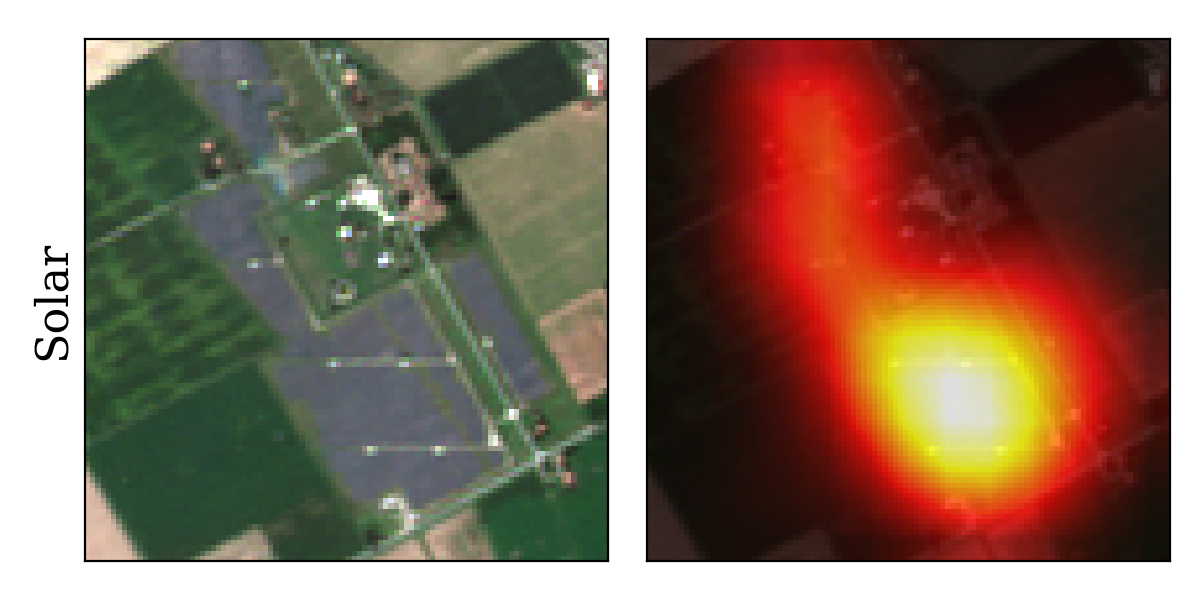}
  \hspace{5em}
  \includegraphics[width=0.35\linewidth]{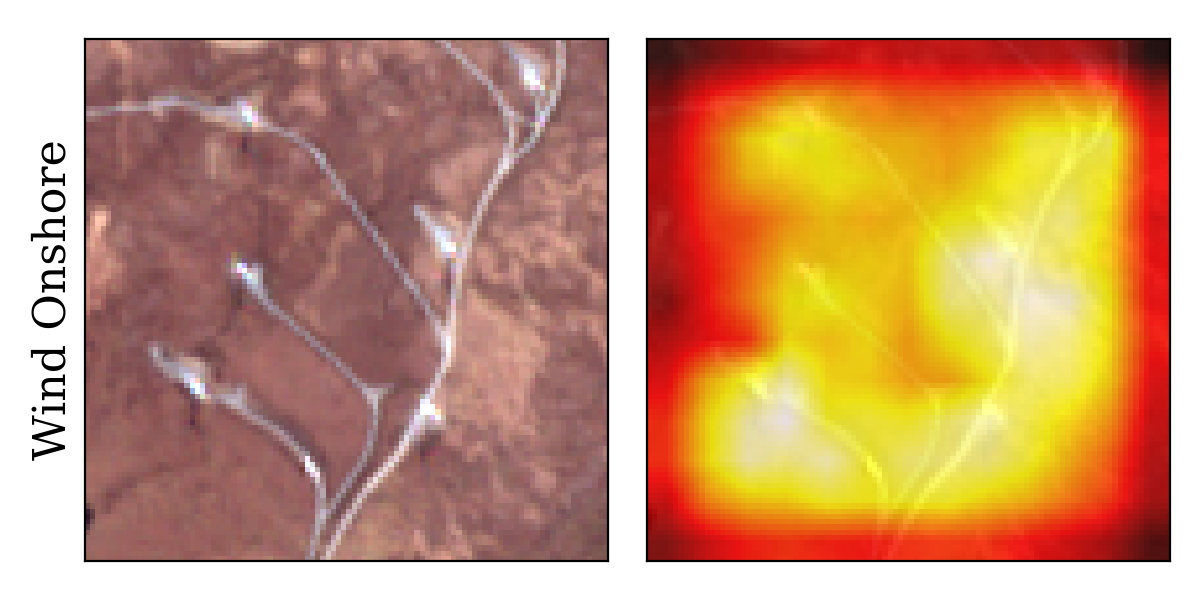}
  \includegraphics[width=0.35\linewidth]{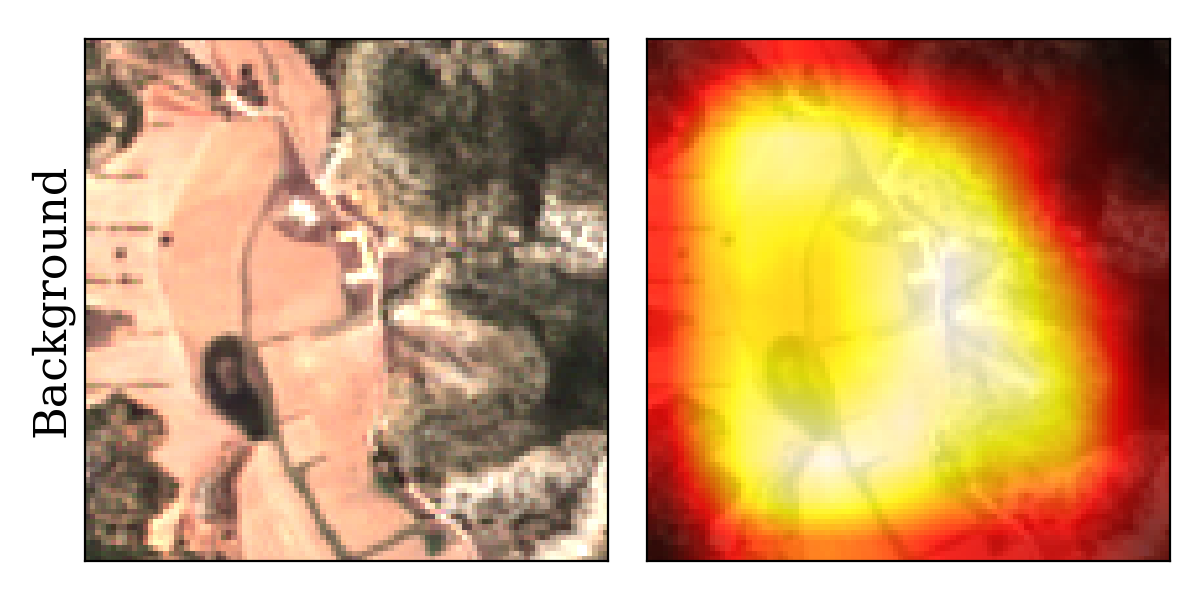}
  \hspace{5em}
  \includegraphics[width=0.35\linewidth]{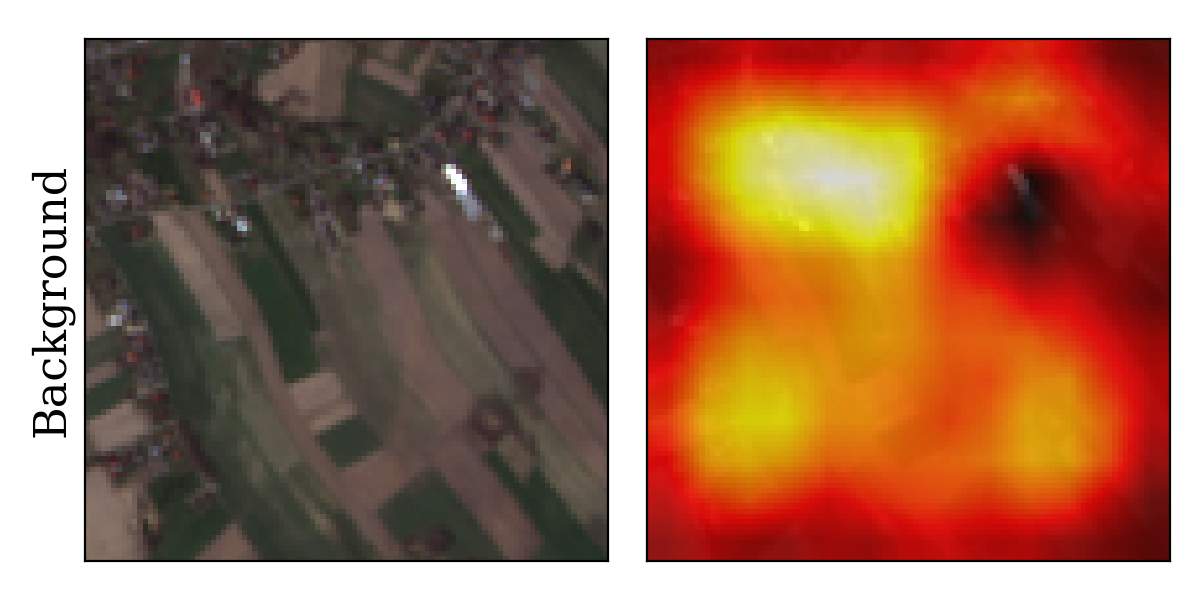}
  \caption{Image examples for all power plant types and the background
    class. For each image pair, the left image shows an RGB image of
    the site, while the right image shows the Class Activation Map
    (CAM, see Section \ref{sec:discussion}) of the corresponding
    class. All examples shown here have been identified correctly by
    our model (see Section \ref{sec:results_pp}). In most cases, the
    most important image regions for the classification (brightest
    region in the CAM) correlate well with characteristic image
    features of the corresponding power plant type.
  \label{fig:examples}}
\end{figure}

In order to verify that predictions made by our architecture are based
on salient features relevant to the classification task, we utilize
Class Activation Maps (CAM, \cite{Zhou2015}). CAMs highlight the most
important image areas in the model's classification process. Examples,
shown in Figure \ref{fig:examples}, show that CAMs indeed focus on
relevant image regions. For instance, coal-powered plants are
identified based on the presence of coal piles, oil and gas power
plants based on storage tanks, solar plants based on their unique
shape and spectral behavior, and hydroelectric plants based on the
presence of gorges, water bodies, and dam structures. In the case of
the background class, activation maps are typically more uniform than
for power plant images due to the lack of characteristic features.

We would like to point out that our results are based on a spatially
limited sample of power plants (all located in Europe), which is
likely to have a beneficial effect on our classification
accuracies. For instance, brown coal power plants in our sample are
likely to be located in Eastern Europe, whereas hydroelectric plants
are mainly located in Scandinavia or the Alps. This localization and
its effect on site surroundings, but also similarities in the design
of plants, are likely to be learned by our model and are thus
reflected by its results.

Nevertheless, the results presented in this work underline the
suitability of medium-resolution remote imaging data to classify
different power plant types, providing the ability to estimate the
energy mix in a given region. We conclude that extending this
classification method to other industrial sites would be worthwhile
and strongly improve the value of land use classification information.

\bibliographystyle{ieeetr}
\bibliography{refs}

\end{document}